\documentclass[letterpaper,10pt,conference]{ieeeconf}
\IEEEoverridecommandlockouts%
\usepackage{cite}
\usepackage{amsmath,amssymb,amsfonts}
\usepackage{algorithmic}
\usepackage{graphicx}
\usepackage{capt-of}
\usepackage{textcomp}
\usepackage{xcolor}
\usepackage{booktabs}
\usepackage{multirow}
\usepackage{url}
\usepackage[hidelinks]{hyperref}
\def\BibTeX{{\rm B\kern-.05em{\sc i\kern-.025em b}\kern-.08em
    T\kern-.1667em\lower.7ex\hbox{E}\kern-.125emX}}
\begin{document}

\title{\LARGE\bfseries LieSpline-DP: Lie-Group B-Spline Diffusion Policy 
\\for Smooth Robot Manipulation}

\author{Erxuan Xie$^{1*}$, Bang Liu$^{1*}$, Pingyun Nie$^{1}$,
    Xingkai Liu$^{1}$, Zhuang Fu$^{1}$, and Bo Zhang$^{1\dagger}$%
\thanks{$^{*}$Equal contribution. $^{\dagger}$Corresponding author.}%
\thanks{$^{1}$All authors are with Shanghai Jiao Tong University.}%
}

\IEEEaftertitletext{%
\begin{minipage}{\textwidth}
\centering
\includegraphics[width=0.98\textwidth]{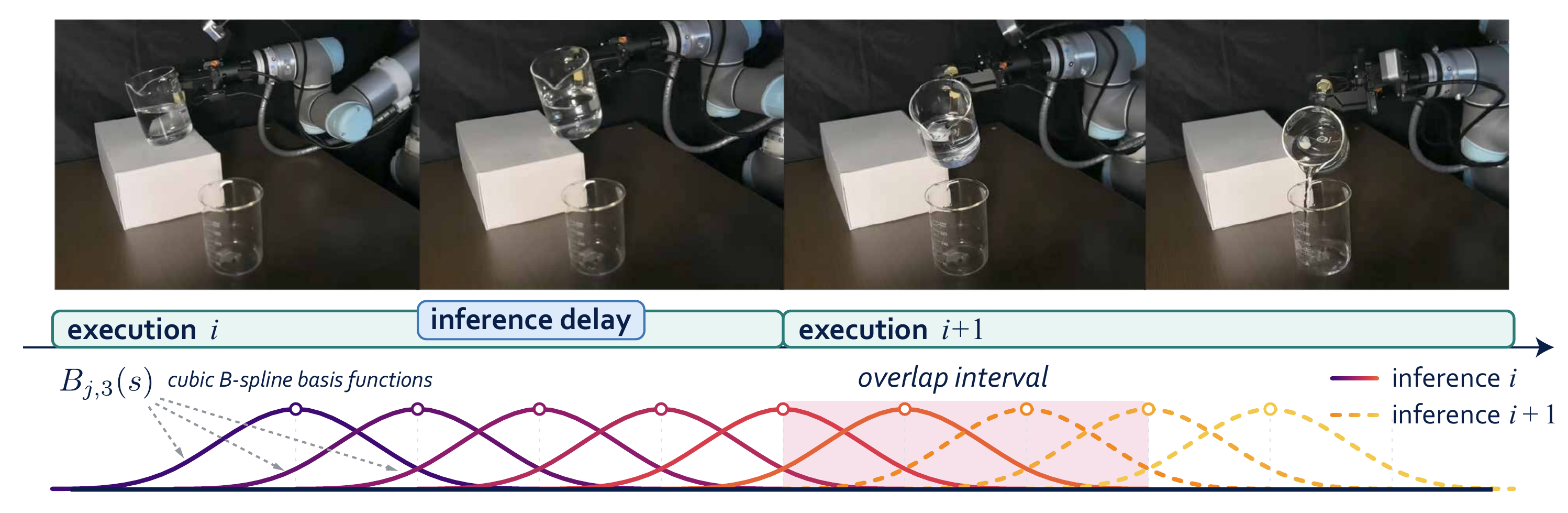}
\captionof{figure}{\textbf{Asynchronous execution and spline extension in LieSpline-DP.} Top: snapshots of a pouring task. Bottom: inference $i+1$ overlaps in time with execution $i$ (blue interval). Solid and dashed curves denote cubic B-spline basis functions associated with controls from inference $i$ and $i+1$, respectively. $B_{j,3}(s)$ denotes the $j$th cubic basis function evaluated at normalized spline phase $s$. The pink overlap interval marks the transition early in execution $i+1$, where inherited and newly predicted controls jointly influence the trajectory.}
\label{fig:inference_timeline}
\vspace{0.5em}
\end{minipage}
}

\maketitle
\thispagestyle{empty}
\pagestyle{empty}

\begin{abstract}
% !!!! FOR ICRA !!!!
% Diffusion Policy (DP) is a powerful Learning from Demonstration (LfD) method for robotic manipulation, yet it suffers from discontinuous and non-smooth trajectories. Spline-based action representations promote smooth motion within individual action chunks, but existing spline-based methods neither guarantee cross-chunk $C^2$ continuity nor account for the group structure of $\mathrm{SE}(3)$. We therefore propose LieSpline-DP, a Lie-group B-spline diffusion policy that generates end-effector trajectories directly on $\mathrm{SE}(3)$ and couples consecutive plans by sharing their boundary control poses, ensuring $C^2$ continuity throughout the entire planned trajectory. Across three real-robot tasks, LieSpline-DP produces lower trajectory jerk and higher task success rates than the DP baseline. The gains are particularly pronounced in real-world tasks involving liquids and flexible objects: in our real-robot experiments, LieSpline-DP achieved a 100\% success rate on both \textit{pouring} and \textit{bucket hooking}, whereas the DP baseline achieved only 10\% and 30\%, respectively. An anonymized version of our code is available at \url{https://anonymous.4open.science/r/LieSpline-DP-775F}.

% !!!! FOR ARXIV !!!!
Diffusion Policy (DP) is a powerful Learning from Demonstration (LfD) method for robotic manipulation, yet it suffers from discontinuous and non-smooth trajectories. Spline-based action representations promote smooth motion within individual action chunks, but existing spline-based methods neither guarantee cross-chunk $C^2$ continuity nor account for the group structure of $\mathrm{SE}(3)$. We therefore propose LieSpline-DP, a Lie-group B-spline diffusion policy that generates end-effector trajectories directly on $\mathrm{SE}(3)$ and couples consecutive plans by sharing their boundary control poses, ensuring $C^2$ continuity throughout the entire planned trajectory. Across three real-robot tasks, LieSpline-DP produces lower trajectory jerk and higher task success rates than the DP baseline. The gains are particularly pronounced in real-world tasks involving liquids and flexible objects: in our real-robot experiments, LieSpline-DP achieved a 100\% success rate on both \textit{pouring} and \textit{bucket hooking}, whereas the DP baseline achieved only 10\% and 30\%, respectively. More details are available at \url{https://xieerxuan.github.io/LieSpline-DP/}.
\end{abstract}

\section{Introduction}

Learning visuomotor policies from demonstrations is an effective paradigm for robotic manipulation. Diffusion Policy (DP), in particular, models multimodal action distributions through conditional denoising and has achieved strong performance in simulated and real-world tasks~\cite{ho2020ddpm,chi2023diffusion}. DP typically predicts a fixed-horizon action chunk, executes part of it, and then replans. On a physical robot, non-negligible inference latency motivates predicting the next chunk while the current one is still being executed. Switching between independently generated chunks, however, can introduce discontinuities and jerky movements, leading to tracking errors and even task failure. Action Chunking with Transformers (ACT) mitigates this problem through output-level temporal ensembling, which takes a weighted average of the actions predicted by overlapping chunks; however, such averaging may mix and blur distinct action modes~\cite{zhao2023act}. Stateful diffusion instead conditions each new prediction on prior actions~\cite{liu2024diffcontrol}. Although this strategy improves cross-chunk coherence, its coupling is learned and does not explicitly preserve the action prefix committed during asynchronous inference. For asynchronous execution, Real-Time Chunking (RTC) freezes the committed prefix and inpaints the remaining actions at test time~\cite{black2025rtc}. Later approaches internalize prefix conditioning during training or reshape the generative flow to learn continuation dynamics~\cite{black2025training,liu2026legato}. However, these methods all operate on discrete actions and do not explicitly guarantee cross-chunk velocity or acceleration continuity.

\begin{table*}[t]
\centering
\caption{Comparison of action representations and execution mechanisms. Existing methods parameterize end-effector or joint actions in Euclidean vector spaces, whereas LieSpline-DP constructs end-effector trajectories directly on $\mathrm{SE}(3)$.}
\label{tab:method_comparison}
\footnotesize
\setlength{\tabcolsep}{6pt}
\renewcommand{\arraystretch}{1.1}
\begin{tabular*}{\linewidth}{@{\extracolsep{\fill}}lccc@{}}
\toprule
Method
& Trajectory Representation
& Cross-Chunk Coupling
& Continuity Property \\
\midrule

DP~\cite{chi2023diffusion}
& Dense waypoints in $\mathbb{R}^d$
& None
& No explicit guarantee \\

ACT~\cite{zhao2023act}
& Dense waypoints in $\mathbb{R}^d$
& Temporal ensembling
& No explicit guarantee \\

RTC~\cite{black2025rtc}
& Dense waypoints in $\mathbb{R}^d$
& Prefix inpainting
& No explicit guarantee \\

BEAST~\cite{zhou2025beast}
& Euclidean B-spline
& Endpoint anchoring
& Within-chunk $C^2$; cross-chunk $C^0$ \\

Spline Policy~\cite{tian2026spline}
& Euclidean Bernstein spline
& Continuity constraints
& Within-chunk $C^2$; cross-chunk $C^0/C^1$ \\

B-spline Policy~\cite{han2026bspline}
& Euclidean B-spline
& Segment alignment
& Within-chunk $C^2$ \\

ABPolicy~\cite{yang2026abpolicy}
& Euclidean B-spline
& Constrained refitting
& Approximately Cross-chunk $C^2$ \\

\textbf{LieSpline-DP}
& \textbf{Lie-group B-spline on $\mathrm{SE}(3)$}
& \textbf{Control-pose prefix inpainting}
& \textbf{Cross-chunk $C^2$} \\

\bottomrule
\end{tabular*}
\end{table*}

The intrinsic smoothness of splines has recently motivated their use as compact action representations~\cite{carvalho2024mpd,zhou2025beast,tian2026spline,han2026bspline,yang2026abpolicy}. A cubic B-spline is $C^2$-continuous within a predicted chunk~\cite{deboor2001splines}, but this property does not automatically extend across independently generated chunks. Existing methods adopt different mechanisms to couple consecutive chunks. For example, BEAST anchors the first control point of a new clamped B-spline to the final action of the previous chunk, guaranteeing positional ($C^0$) continuity without ensuring matching velocities or accelerations at the boundary~\cite{zhou2025beast}. ABPolicy instead refits a spline over past and future actions using least squares, improving boundary smoothness but only approximately matching the executed trajectory~\cite{yang2026abpolicy}. Although these mechanisms improve continuity between consecutive chunks, existing spline-based methods still lack a formal guarantee of cross-chunk $C^2$ continuity.

The methods mentioned above also share a geometric limitation: they parameterize joint actions or end-effector poses in Euclidean vector spaces. For rigid-body motion, Euclidean denoising and interpolation do not intrinsically respect the group structure of $\mathrm{SE}(3)$; the resulting rotational trajectories and their smoothness therefore depend on the chosen rotation representation~\cite{sola2018micro}. Recent work incorporates rigid-motion geometry into diffusion-based manipulation through diffusion-derived cost fields and bi-equivariant pose distributions on $\mathrm{SE}(3)$ for grasp and target-pose generation~\cite{urain2023se3diffusionfields,ryu2024diffusionedfs}. Other approaches extend this direction to closed-loop policies that generate action sequences equivariant to $\mathrm{SE}(3)$ or the broader $\mathrm{SIM}(3)$ group~\cite{yang2025equibot,tie2025etseed,zhu2025sdp}. These methods, however, primarily address spatial geometry rather than action smoothness.

Motivated by these observations, we propose \textbf{LieSpline-DP}, a Lie-group B-spline diffusion policy that generates end-effector trajectories directly on $\mathrm{SE}(3)$. The policy predicts compact six-dimensional local coordinates of $\mathrm{SE}(3)$ control poses, which a cumulative Lie-group B-spline decodes through group composition into a continuous rigid-body trajectory~\cite{kim1995quaternion,sommer2020efficient}. As illustrated in Fig.~\ref{fig:inference_timeline}, successive plans are connected through a transition interval early in execution $i+1$, where the basis functions associated with control poses from inference $i$ (solid) and $i+1$ (dashed) overlap. By reusing boundary controls, LieSpline-DP extends the trajectory as a single $C^2$-continuous Lie-group B-spline, as detailed in Section~\ref{sec:asynchronous_pipeline}.

To the best of our knowledge, LieSpline-DP is the first closed-loop visuomotor diffusion policy for general-purpose manipulation to use a B-spline defined directly on $\mathrm{SE}(3)$ as its end-effector action representation. Table~\ref{tab:method_comparison} compares LieSpline-DP with prior spline-based generative policies and other related baseline policies in action geometry, cross-chunk coupling, and continuity guarantees.

Our main contributions are summarized as follows:
\begin{itemize}
    \item \textbf{Lie-group B-spline action representation.}
    We formulate visuomotor policy learning as conditional diffusion over local coordinates of $\mathrm{SE}(3)$ control poses rather than dense waypoints. A cumulative Lie-group B-spline decodes these compact controls through group composition, defining the continuous end-effector trajectory directly on $\mathrm{SE}(3)$.

    \item \textbf{Asynchronous replanning with cross-chunk $C^2$ continuity.}
    We couple consecutive plans by fixing an inherited control-pose prefix and inpainting only future controls. With a shared uniform knot lattice and aligned physical timing, reusing three boundary control poses guarantees cross-chunk $C^2$ continuity of the cubic spline reference, without post-hoc blending or refitting.

    \item \textbf{Execution-aligned spline training data construction.}
    We construct paired control-pose prefixes and future targets from dense demonstrations using phase-offset Lie-group spline fitting. Event weighting emphasizes critical waypoints, while latency-aware windowing aligns observation histories with future execution boundaries, providing training samples tailored to the policy's asynchronous continuation mechanism.
\end{itemize}

\section{Preliminaries}
\label{sec:preliminaries}

\begin{figure*}[t]
\centering
\includegraphics[width=\textwidth]{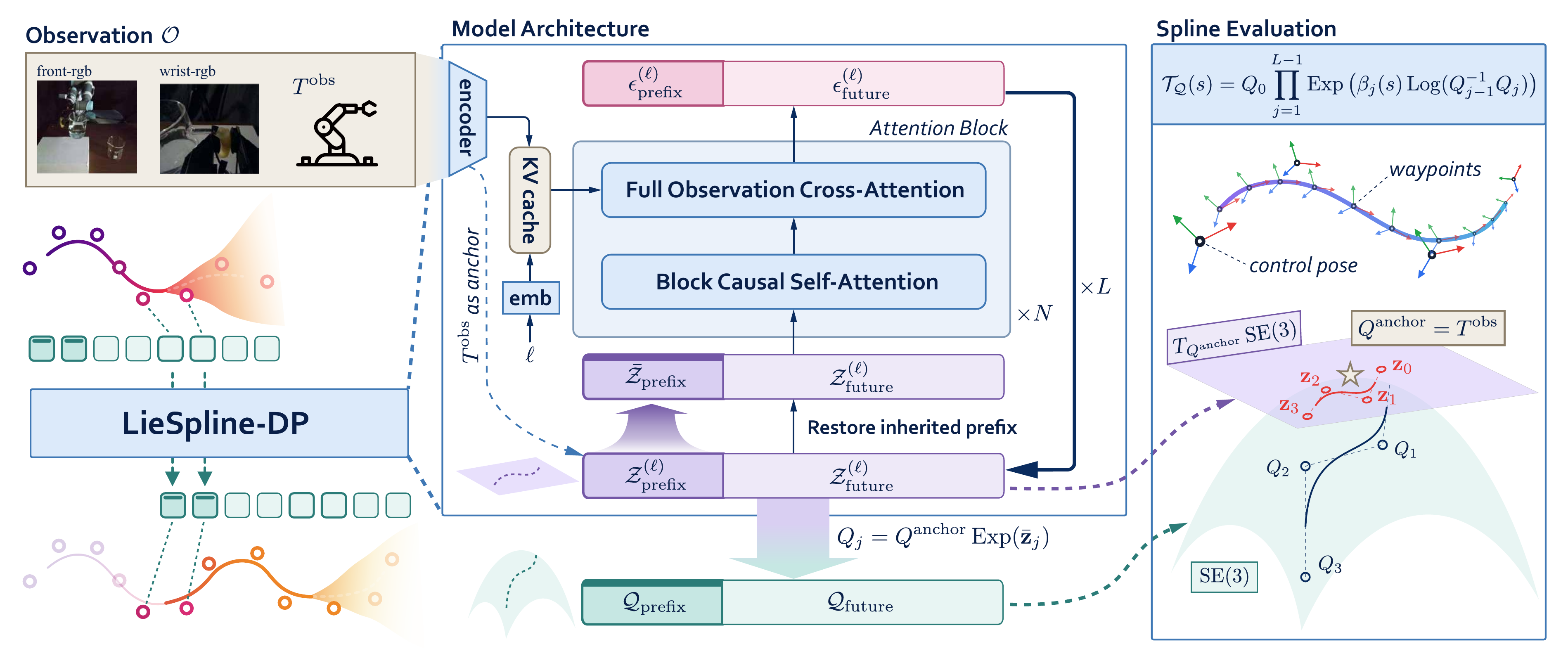}
\caption{\textbf{LieSpline-DP architecture.} An observation encoder conditions a diffusion model over local control-pose coordinates. Block-causal self-attention separates the fixed boundary prefix from the denoised future tokens. The resulting coordinates are lifted to $\mathrm{SE}(3)$ and decoded by a cumulative Lie-group B-spline into dense end-effector waypoints.}
\label{fig:model_architecture}
\end{figure*}

\subsection{Diffusion Policy}
Diffusion Policy models a conditional distribution over action sequences by reversing a gradual noising process~\cite{chi2023diffusion}. Let $\mathbf x^{(0)}$ denote a clean action representation and let $\ell$ index the diffusion step. The forward process admits the closed form
\begin{equation}
    \mathbf x^{(\ell)}=
    \sqrt{\bar\alpha_\ell}\,\mathbf x^{(0)}
    +\sqrt{1-\bar\alpha_\ell}\,\boldsymbol\epsilon,
    \qquad \boldsymbol\epsilon\sim\mathcal N(0,I),
    \label{eq:forward_diffusion}
\end{equation}
where $\bar\alpha_\ell$ is determined by a prescribed noise schedule. A denoising network $\boldsymbol\epsilon_\theta(\mathbf x^{(\ell)},\ell,\mathcal O)$ is trained to recover the injected noise from the noisy action and observation context $\mathcal O$:
\begin{equation}
    \mathcal L_{\mathrm{diff}}
    =\mathbb E_{\mathbf x^{(0)},\ell,\boldsymbol\epsilon}
    \left[\left\|\boldsymbol\epsilon-
    \boldsymbol\epsilon_\theta(\mathbf x^{(\ell)},\ell,\mathcal O)
    \right\|_2^2\right].
    \label{eq:diffusion_objective}
\end{equation}
At inference time, the learned noise predictor defines a sequence of reverse transitions that transforms Gaussian noise into an action sequence conditioned on the observation. Known action entries can be enforced throughout this process by overwriting the corresponding generated entries after each denoising step, thereby enabling conditional generation via inpainting~\cite{lugmayr2022repaint}. Sampling can be accelerated with Denoising Diffusion Implicit Models (DDIM), which support generation using fewer reverse denoising steps~\cite{song2021ddim}.

\subsection{Geometry of \texorpdfstring{$\mathrm{SE}(3)$}{SE(3)} B-spline Trajectories}
An end-effector pose is represented by
\begin{equation}
    T=\begin{bmatrix}R&\mathbf t\\\mathbf 0^\top&1\end{bmatrix}
    \in\mathrm{SE}(3),
    \qquad R\in\mathrm{SO}(3),\quad \mathbf t\in\mathbb R^3.
\end{equation}
The exponential and logarithm maps relate a pose increment to a twist $\boldsymbol\xi=[\boldsymbol\rho^\top,\boldsymbol\phi^\top]^\top\in\mathbb R^6$ through
\begin{equation}
    T_b=T_a\operatorname{Exp}(\boldsymbol\xi),\qquad
    \boldsymbol\xi=\operatorname{Log}(T_a^{-1}T_b).
    \label{eq:se3_increment}
\end{equation}
Throughout this work, we adopt this right-multiplicative increment convention: the pose increment acts on the right of the reference pose. These maps provide local Euclidean coordinates while preserving group composition in the decoded trajectory.

Let $\mathcal Q=(Q_0,\ldots,Q_{H-1})$ be a sequence of control poses and define the relative twists $\boldsymbol\Omega_j=\operatorname{Log}(Q_{j-1}^{-1}Q_j)$. The corresponding cumulative Lie-group B-spline is~\cite{sommer2020efficient}
\begin{equation}
    \mathcal T_{\mathcal Q}(s)
    =Q_0\prod_{j=1}^{H-1}
    \operatorname{Exp}\!\left(\beta_{j,p}(s)\boldsymbol\Omega_j\right),
    \label{eq:lie_spline}
\end{equation}
where factors are composed in increasing order of $j$. Here, $\beta_{j,p}(s)=\sum_{i=j}^{H-1}B_{i,p}(s)$ is the cumulative basis function constructed from the degree-$p$ B-spline basis functions $B_{i,p}(s)$. Throughout this work, we use the uniform knot vector $\boldsymbol\tau=\{\tau_i\}$ in the normalized spline phase $s$, with unit-spaced knots $\tau_i=i$. Under this uniform construction, the B-spline basis functions $B_{j,p}(s)$ have support on $s\in[j,j+p+1)$. The lower plot in Fig.~\ref{fig:inference_timeline} shows the cubic B-spline basis functions associated with this uniform knot vector.

\section{Method: LieSpline-DP}
\label{sec:method}

\subsection{Overview}

The core contribution of LieSpline-DP lies in its action representation and cross-plan coupling mechanism, and is therefore compatible with different DP backbones. In this work, we instantiate it with a Transformer-based diffusion denoiser~\cite{vaswani2017attention,peebles2023dit}.

At replanning cycle $n$, LieSpline-DP represents the end-effector trajectory as a degree-$p$ B-spline on $\mathrm{SE}(3)$, using $H=K+F$ control poses, denoted by $\mathcal Q_n=(Q_{n,0},\ldots,Q_{n,H-1})$. As shown in Fig.~\ref{fig:model_architecture}, the observation encoder first encodes the observation history $\mathcal O_n$ into an observation memory stored as a key--value (KV) cache. The observed end-effector pose $T_n^{\mathrm{obs}}$, which is contained in the robot-state observation, defines the anchor pose $Q_n^{\mathrm{anchor}}=T_n^{\mathrm{obs}}$ of a local tangent-space chart. The control poses are represented in this chart by $\mathcal Z_n=(\mathbf z_{n,0},\ldots,\mathbf z_{n,H-1})$. Its first $K=p$ entries form a fixed prefix inherited from the tail control poses of the preceding execution, shown with an overbar on the token symbols in Fig.~\ref{fig:model_architecture}; the remaining $F$ entries represent the future control poses to be generated. The diffusion model therefore operates as a control-point inpainting model, with block-causal self-attention allowing the noisy future tokens to attend to the fixed prefix and cross-attention supplying the observation memory.

The completed tangent-vector sequence is then lifted back to absolute control poses on $\mathrm{SE}(3)$ through
\begin{equation}
    Q_{n,j}=Q_n^{\mathrm{anchor}}\operatorname{Exp}(\mathbf z_{n,j}),
    \label{eq:lift_control_pose}
\end{equation}
for $j=0,\ldots,H-1$. These poses define the cumulative Lie-group B-spline shown in the rightmost panel of Fig.~\ref{fig:model_architecture}. The lower-right illustration summarizes the geometric relationship among the local coordinates $\mathcal Z_n \in T_{Q^\mathrm{anchor}} \mathrm{SE}(3)$, the control poses $\mathcal Q_n \in \mathrm{SE}(3)$, and the resulting continuous trajectory.

\subsection{Asynchronous Inference Pipeline}
\label{sec:asynchronous_pipeline}

\textbf{Local $\mathrm{SE}(3)$ control poses.}
The observation history contains the measured end-effector pose, which defines the anchor $Q_n^{\mathrm{anchor}}=T_n^{\mathrm{obs}}\in\mathrm{SE}(3)$. We use this pose both as policy input and as the origin of the local chart in which diffusion is performed. For $j=0,\ldots,H-1$, the local and absolute representations are related by
\begin{equation}
    \mathbf z_{n,j}=\operatorname{Log}\!\left((Q_n^{\mathrm{anchor}})^{-1}Q_{n,j}\right).
    \label{eq:local_control_pose}
\end{equation}
Here $\mathbf z_{n,j}=[\boldsymbol\rho_{n,j}^{\top},\boldsymbol\phi_{n,j}^{\top}]^{\top}\in\mathbb R^6$ follows the right-increment convention in~\eqref{eq:se3_increment}. Because $Q_n^{\mathrm{anchor}}$ is close to the trajectory being predicted, these coordinates provide a well-conditioned local linearization for Gaussian noising and denoising. At each inference, the inherited control-pose prefix is stored as absolute poses on $\mathrm{SE}(3)$ and re-expressed relative to the newly observed anchor at each cycle. Consequently, changing the anchor alters only the network input coordinates, leaving the physical boundary trajectory unchanged.

\textbf{Fixed-prefix inpainting diffusion.}
At cycle $n$, the inherited poses are first mapped to local coordinates using the current anchor $Q_n^{\mathrm{anchor}}$. Let $\tilde{\mathbf z}_{n,j}$ denote their normalized coordinates. They occupy the first $K$ tokens of the diffusion sequence, while the future tokens are initialized with Gaussian noise. The Transformer denoiser first applies block-causal self-attention with an asymmetric mask: prefix tokens attend only to the prefix, whereas future tokens attend to both the prefix and the future block. It then uses cross-attention to condition the resulting token representations on the visual--proprioceptive memory produced by the observation encoder; the corresponding key and value projections are stored in a KV cache and reused across reverse-denoising steps.

The inherited prefix is restored after every forward-noising or reverse-denoising update:
\begin{equation}
    \tilde{\mathbf z}_{n,j}^{(\ell)}
    \leftarrow\tilde{\mathbf z}_{n,j}^{(0)},
    \qquad j=0,\ldots,K-1.
    \label{eq:hard_prefix}
\end{equation}
Consequently, sampling cannot modify the boundary controls committed by the previous plan.

\textbf{Asynchronous execution with $C^2$ continuity.}
As illustrated by the basis functions in Fig.~\ref{fig:inference_timeline}, the smoothness guarantee relies on two coupled design choices: a simple uniform knot lattice shared by all replanning cycles, and the reuse of boundary control poses. With no repeated interior knots, a degree-$p$ B-spline is $C^{p-1}$-continuous within each plan~\cite{deboor1972calculating}. The cumulative Lie-group B-spline in~\eqref{eq:lie_spline} inherits this $C^{p-1}$ continuity as a pose curve on $\mathrm{SE}(3)$~\cite{kim1995quaternion,sommer2020efficient}. For cubic splines under the uniform execution timing used here, this means that the reference pose, its body-frame twist (linear and angular velocity), and the time derivative of that twist are all continuous within each plan. Across two consecutive plans $n$ and $n+1$, the common knot spacing and phase convention align the handover at phase $s=p+E$ of plan $n$ with phase $s=p$ of plan $n+1$. The local bases are thus exact translates over equal physical knot spans, and once plan $n$ advances by $E$ control intervals, plan $n+1$ reuses the $K=p$ control poses at the handover:
\begin{equation}
    Q_{n+1,j}=Q_{n,E+j},\qquad j=0,\ldots,K-1,
    \qquad K=p.
    \label{eq:shared_controls}
\end{equation}
These overlapping poses form the fixed prefix of the new control lattice. The remaining $F$ control poses are generated from the latest available observation. For the initial plan, where no preceding lattice exists, we use the stationary prefix
\begin{equation}
    Q_{0,j}=Q_0^{\mathrm{anchor}},\qquad j=0,\ldots,K-1.
    \label{eq:initial_controls}
\end{equation}

After denoising, the predicted future coordinates are lifted to absolute control poses using~\eqref{eq:lift_control_pose}. The resulting poses are concatenated with the inherited absolute prefix and decoded by the cumulative Lie-group spline in~\eqref{eq:lie_spline}. We sample this continuous curve at the controller rate and append the resulting $W$ reference poses to the execution queue:
\begin{equation}
    T_{n,k}=\mathcal T_{\mathcal Q_n}\!\left(p+\frac{k}{S}\right),
    \qquad k=1,\ldots,W,\qquad W=ES.
    \label{eq:execution_phase}
\end{equation}
Here $S$ is the number of controller steps per control interval. As shown in Fig.~\ref{fig:inference_timeline}, inference for next plan starts a few steps before handover so that computation overlaps execution.

Under this aligned-knot, shared-prefix construction, consecutive plans define a single growing cubic Lie-group B-spline that is $C^2$-continuous throughout. For robot execution, waypoints are sampled from this reference spline.

\begin{figure}[t]
\centering
\includegraphics[width=0.90\columnwidth]{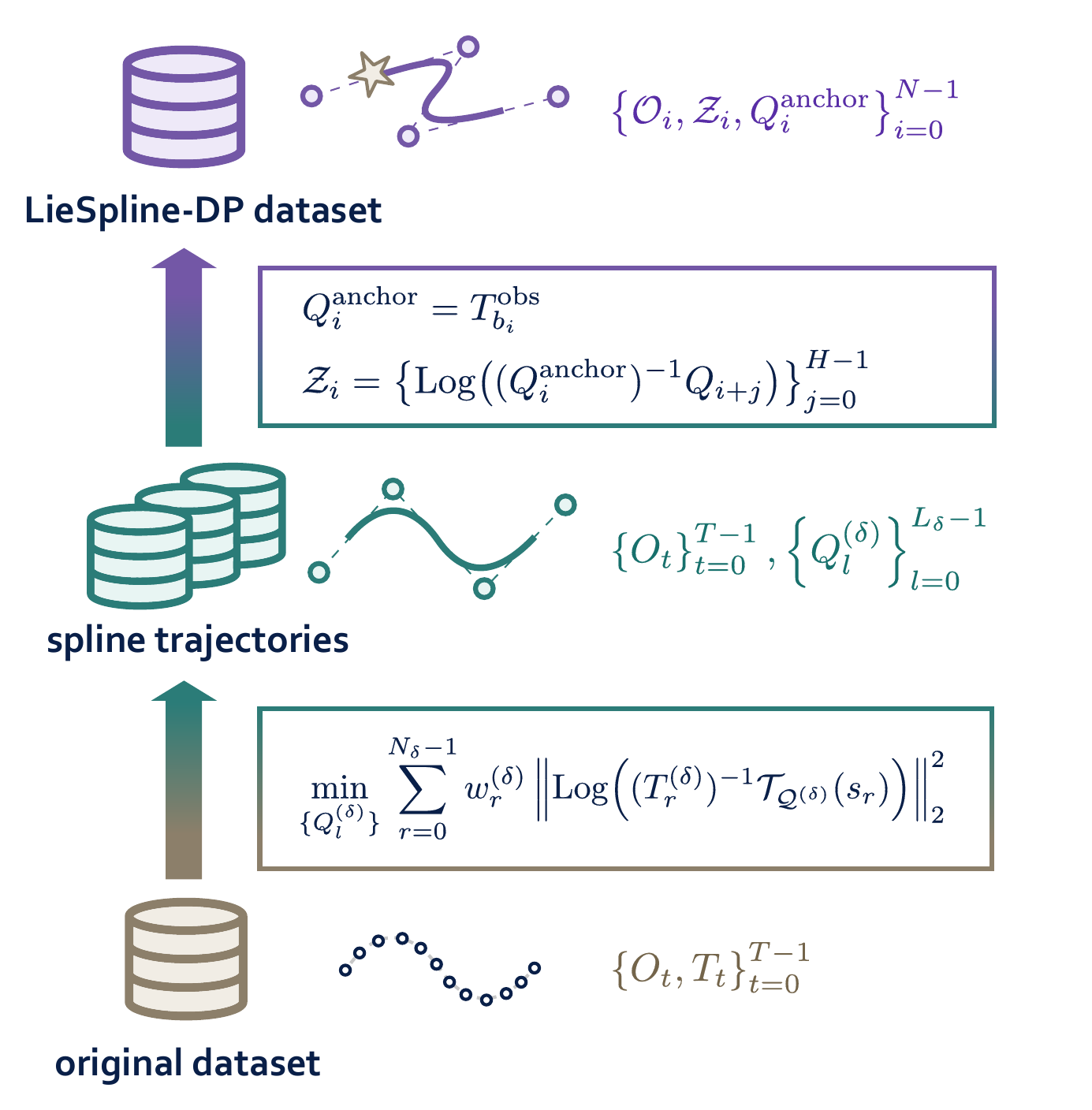}
\caption{\textbf{Offline dataset conversion pipeline.} From bottom to top, each dense demonstration is converted into phase-offset Lie-group spline trajectories, which are then sliced into observation-conditioned windows of local control-pose coordinates. The displayed anchor corresponds to the synchronous case $L=0$; latency-aligned real-robot windows use the generalized anchor in~\eqref{eq:latency_alignment}.}
\label{fig:dataset_conversion}
\end{figure}

\subsection{Spline Dataset Conversion and Training}

As illustrated in Fig.~\ref{fig:dataset_conversion}, the offline conversion proceeds in three stages, from bottom to top: the original dense demonstration, the fitted $\mathrm{SE}(3)$ spline trajectories, and the resulting LieSpline-DP training dataset. For clarity, the figure depicts only the pose stream; gripper commands are processed using the same temporal alignment.

\textbf{Dense demonstrations and start-offset augmentation.}
An original episode in the bottom row of Fig.~\ref{fig:dataset_conversion} is represented by $\{(\mathcal O_t,T_t^\star)\}_{t=0}^{T-1}$. Because adjacent spline controls are separated by $S$ dense steps, using only one control-lattice phase would retain only approximately one out of every $S$ possible window alignments. We therefore create one sub-episode for each start offset $\delta\in\{0,\ldots,S-1\}$. For offset $\delta$ and $r=0,\ldots,N_\delta-1$, define
\begin{equation}
    (\mathcal O_r^{(\delta)},T_r^{(\delta)},g_r^{(\delta)})
    = (\mathcal O_{\delta+r},T_{\delta+r}^\star,g_{\delta+r}^\star).
    \label{eq:offset_subepisode}
\end{equation}
To make the phase lattice align exactly with the discrete timing of the original waypoints trajectory, each sub-episode is padded by repeating its terminal sample until its length is a multiple of $S$; the repeated samples define a stationary terminal hold.

\textbf{Phase-offset spline trajectories.}
The middle row of Fig.~\ref{fig:dataset_conversion} represents each sub-episode by an absolute control-pose sequence $\mathcal Q^{(\delta)}=\{Q_l^{(\delta)}\}_{l=0}^{L_\delta-1}$. At dense index $r$, the corresponding spline phase is $s_r=p+(r+1)/S$. The control poses are obtained by solving the event-weighted fitting problem:
\begin{equation}
    \min_{\{Q_l^{(\delta)}\}}
    \sum_{r=0}^{N_\delta-1} w_r^{(\delta)}
    \left\|\operatorname{Log}\!\left(
    (T_r^{(\delta)})^{-1}
    \mathcal T_{\mathcal Q^{(\delta)}}(s_r)
    \right)\right\|_2^2,
    \label{eq:spline_fitting}
\end{equation}
Here, $\mathcal T_{\mathcal Q^{(\delta)}}(s_r)$ is evaluated using the cumulative Lie-group B-spline defined in~\eqref{eq:lie_spline}. The six components of the logarithmic residual are standardized by the dataset translation and rotation scales before evaluating the norm. The first $p$ controls are fixed to the measured pose preceding the first action, and the last $p$ controls are fixed to the terminal hold. The waypoint weights $w_r^{(\delta)}$ can be determined from gripper-state transitions, tactile signals, or other task events. Larger weights encourage the spline to fit waypoints more accurately at critical moments, while smaller weights elsewhere place greater emphasis on smoothness.

\textbf{Training windows for LieSpline-DP.}
The top arrow and boxed equations in Fig.~\ref{fig:dataset_conversion} show how the fitted trajectories are converted into training samples. After pooling the phase-offset trajectories, we suppress $\delta$ on the selected control sequence for compactness. Let $n$ index a control-boundary window and let $b_n=nS$ denote its first dense action index. In the synchronous case depicted in the figure, the anchor and the $H$ local control-pose labels are
\begin{equation}
    \begin{aligned}
    &Q_n^{\mathrm{anchor}}=T_{b_n}^{\mathrm{obs}},\\
    &\mathcal Z_n=
    \left\{
    \operatorname{Log}\!\left(
    (Q_n^{\mathrm{anchor}})^{-1}Q_{n+j}
    \right)
    \right\}_{j=0}^{H-1}.
    \end{aligned}
    \label{eq:dataset_window}
\end{equation}
Thus the final dataset shown in the top row is
\begin{equation}
    \mathcal D_{\mathrm{LieSpline}}
    =\left\{
    (\mathcal O_n,\mathcal Z_n,Q_n^{\mathrm{anchor}})
    \right\}_{n=0}^{N-1},
    \label{eq:liespline_dataset}
\end{equation}
where $\mathcal O_n$ denotes the observation history ending at $b_n$. The first $K$ elements of $\mathcal Z_n$ are the inherited prefix and the remaining $F$ elements are the future prediction target, with $H=K+F$.

For asynchronous execution, inference is triggered $L$ dense steps before the same action boundary. Specifically,
\begin{equation}
    a_n=\max(b_n-L,0),\qquad
    Q_n^{\mathrm{anchor}}=T_{a_n}^{\mathrm{obs}},
    \label{eq:latency_alignment}
\end{equation}
and $\mathcal O_n$ now ends at $a_n$. Substituting this anchor into~\eqref{eq:dataset_window} gives the real-robot training label, while $L=0$ reduces exactly to the boxed formula in Fig.~\ref{fig:dataset_conversion}. Gripper labels remain aligned to the execution boundary $b_n$; their platform-specific parameterization does not affect the $\mathrm{SE}(3)$ continuity result.

The denoising loss is evaluated only on valid, unconditioned entries,
\begin{equation}
    \mathcal L
    =\mathcal L_{\mathrm{pose}}
    +\lambda_g\mathcal L_{\mathrm{grip}},
    \label{eq:diffusion_loss}
\end{equation}
where padded entries are masked and $\lambda_g$ balances the auxiliary gripper objective. During training, we may perturb the inherited prefix poses to improve robustness to the distribution shift between demonstration-derived prefixes used in training and model-predicted prefixes encountered at inference.

\begin{table*}[t]
\centering
\caption{Robomimic PH low-dimensional results averaged over three training seeds. Success and jerk entries report Best / Last-10. Jerk values are the four-task mean of the task-level 95th percentile (p95).}
\label{tab:simulation_results}
\footnotesize
\setlength{\tabcolsep}{6pt}
\renewcommand{\arraystretch}{1.1}
\begin{tabular*}{\linewidth}{@{\extracolsep{\fill}}lcccccc@{}}
\toprule
& \multicolumn{4}{c}{Success rate (\%, Best / Last-10)} & \multicolumn{2}{c}{Four-task mean jerk p95 (Best / Last-10)} \\
\cmidrule(lr){2-5}\cmidrule(lr){6-7}
Method & Lift & Can & Square & Tool Hang
& Translational ($\mathrm{m}/\mathrm{s}^3$) $\downarrow$
& Rotational ($\mathrm{rad}/\mathrm{s}^3$) $\downarrow$ \\
\midrule
DP
& 100.00 / 99.80
& 100.00 / 99.20
& 99.33 / 91.60
& 97.33 / 86.87
& 59.13 / 50.46 & 140.50 / 120.34 \\
\textbf{LieSpline-DP}
& 100.00 / 99.93
& 100.00 / 98.87
& 97.33 / 89.07
& 93.33 / 81.27
& \textbf{11.57 / 9.73}
& \textbf{19.83 / 17.11} \\
\bottomrule
\end{tabular*}
\end{table*}

\begin{figure*}[t]
\centering
\includegraphics[width=\textwidth]{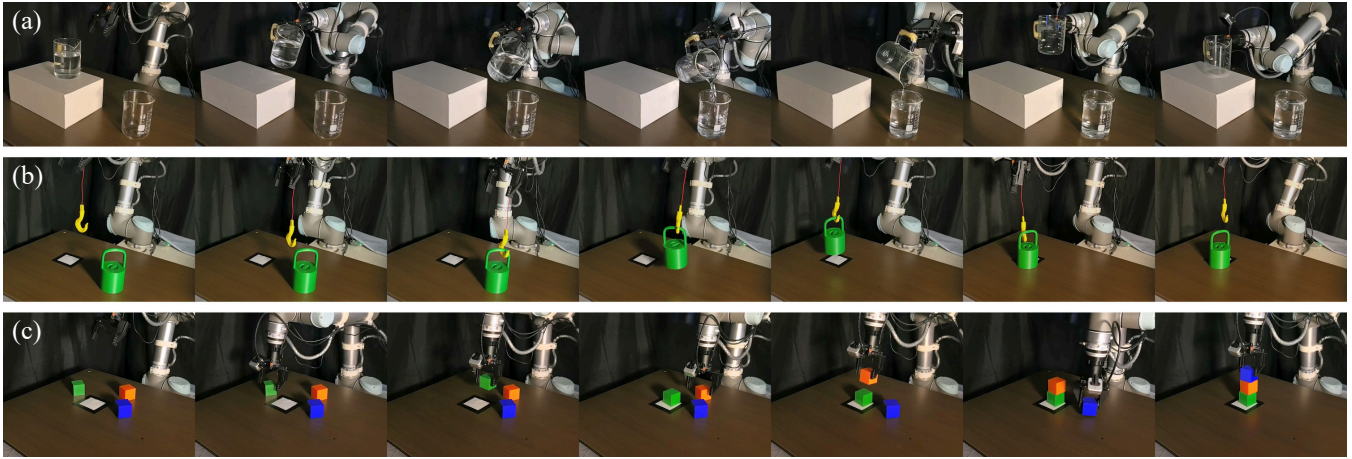}
\caption{\textbf{Keyframe snapshot sequences for the three real-robot tasks.} (a)~Pouring; (b)~Bucket hooking; (c)~Cube stacking.}
\label{fig:real_robot_tasks}
\end{figure*}

\section{Experiments}
\label{sec:experiments}

We evaluate LieSpline-DP with two questions in mind. First, we ask whether replacing dense action chunks with Lie-group spline control poses preserves manipulation performance while improving the smoothness of the executed trajectories. Second, we examine whether these kinematic improvements translate into more reliable behavior on physical tasks. We study the first question through a controlled comparison on four Robomimic benchmarks and the second through three real-robot manipulation tasks.

\subsection{Simulation}

\textbf{Experimental setup.}
We compare LieSpline-DP with dense-action DP~\cite{chi2023diffusion} on the Robomimic Proficient-Human low-dimensional Lift, Can, Square, and Tool Hang benchmarks~\cite{mandlekar2022robomimic}. Both methods share demonstrations, normalization, a Transformer denoiser, diffusion schedule, optimizer, and exponential moving average (EMA) settings. They use absolute end-effector actions and 2 observation steps. Each execution chunk contains 8 waypoints, executed sequentially at $20$~Hz. LieSpline-DP uses $p=K=3$, $F=8$, and $H=11$. We set $S=2$, so consecutive waypoints are separated by $0.5$ units of spline phase, and execute $E=4$ control intervals per replan. Each task uses three training seeds; EMA policies are evaluated every 50 epochs over 50 rollouts with matched environment seeds and initial states.

\textbf{Metrics.}
We report success and executed-trajectory jerk as Best / Last-10: the highest-success checkpoint and the final ten evaluations' average, respectively. Both metrics use the same checkpoints or windows and are averaged over three seeds. Translational jerk uses third-order position differences; rotational jerk uses second-order differences of Lie-log angular velocities. For each task, seed, and evaluation, p95 jerk pools all $20$~Hz samples from the 50 rollouts, including failures. Reported jerk values additionally average across the four tasks.

\textbf{Results.}
Table~\ref{tab:simulation_results} shows four-task mean p95 jerk reductions of 80.4\% / 80.7\% for translation and 85.9\% / 85.8\% for rotation (Best / Last-10). Success rates are similar on Lift and Can, but lower for LieSpline-DP on Square and Tool Hang.

\subsection{Real Robot}

\textbf{Experimental setup and tasks.}
We evaluate on a UR5e arm equipped with a Unitree Dex1 gripper. Policy waypoints are issued at $15$~Hz. For compatibility with standard low-level controllers, we send sampled waypoints rather than spline controls to the robots. The real-robot model uses $p=K=3$, $F=8$, $H=11$, $S=2$, $E=4$, and $W=8$. LieSpline-DP's equivalent execution and prediction horizons, measured in waypoint steps, match those of the DP baseline. Both methods use asynchronous inference, overlapping prediction of the next action chunk with execution of the current one, and are trained on observation--action windows aligned using the same nominal latency of $n_{\mathrm{latency\_steps}}=2$ waypoint steps. We consider the following three tasks:
\begin{itemize}
    \item \textbf{\emph{Pouring}:} grasp a water-filled beaker and pour its contents into a target beaker. A trial succeeds if all water is transferred without spilling.
    \item \textbf{\emph{Bucket hooking}:} use a tethered hook to engage a bucket handle and transport the bucket to a designated target. A trial succeeds if the bucket reaches the target.
    \item \textbf{\emph{Cube stacking}:} stack three randomly scattered cubes at a predefined location in a prescribed color order. A trial succeeds if the three cubes form a stable stack in the prescribed color order at the designated location.
\end{itemize}
Pouring and bucket hooking probe interactions with liquids and flexible objects, respectively. Cube stacking additionally tests precise manipulation. For each task, DP and LieSpline-DP are trained on the same demonstrations: 50 for pouring, 50 for bucket hooking, and 100 for cube stacking including demonstrations of recovery from human interventions. Both methods are trained for the same number of epochs within each task. In the main evaluation, we evaluate each method over 20 trials per task without deliberate human perturbations. We additionally test LieSpline-DP on cube stacking with human perturbations to assess its responsiveness to changes during execution.

\textbf{Metrics.}
We compare task success and the smoothness of the time-stamped TCP target poses actually issued to the robot. Translational jerk is computed by applying three finite differences to the commanded positions using the actual time intervals; angular jerk is obtained from Lie-log angular velocities followed by two finite differences. Derivatives are computed separately within each trial. For each method and task, we then pool the jerk magnitudes over all valid time points from all 20 evaluation trials and report the 95th percentile (p95) and maximum of this pooled distribution, separately for translation and rotation. These metrics characterize commands, not measured TCP motion. Different trajectory signals and sampling rates preclude direct comparison of absolute jerk magnitudes between simulation and hardware.

\begin{table}[t]
\centering
\caption{Success rates on the three real-robot tasks without deliberate human perturbations, with 20 trials per method and task.}
\label{tab:real_robot_results}
\footnotesize
\setlength{\tabcolsep}{6pt}
\renewcommand{\arraystretch}{1.1}
\begin{tabular*}{\linewidth}{@{\extracolsep{\fill}}lccc@{}}
\toprule
Method & Pouring & Bucket hooking & Cube stacking \\
\midrule
DP & 10\% & 30\% & 55\% \\
\textbf{LieSpline-DP} & \textbf{100\%} & \textbf{100\%} & \textbf{80\%} \\
\bottomrule
\end{tabular*}
\end{table}

\begin{figure}[t]
\centering
\IfFileExists{assets/real_robot_commanded_jerk.pdf}{%
    \includegraphics[width=0.8\columnwidth,trim=0bp 5bp 0bp 12bp,clip]{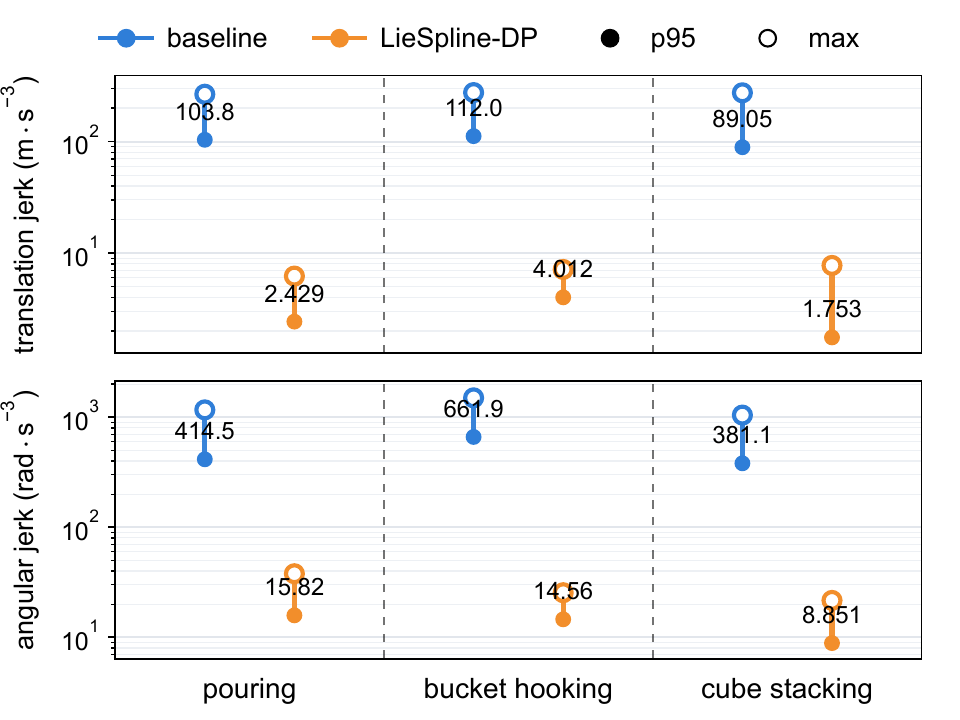}%
}{%
    \fbox{\parbox[c][0.86\columnwidth][c]{0.92\columnwidth}{%
        \centering
        \textbf{Reserved for real-robot commanded-trajectory jerk plot}\\[0.5em]
        Top: translation command jerk\\
        Bottom: angular command jerk\\[0.5em]
        Filled marker: p95; open marker: max\\
        Logarithmic vertical axes
    }}%
}
\caption{\textbf{Jerk comparison for the TCP target trajectories actually issued to the robot.} For each method and task, translational jerk (top) and angular jerk (bottom) magnitudes are pooled over all valid time points from all 20 evaluation trials, with derivatives computed within each trial. Filled and open markers denote the p95 and maximum of the pooled samples, respectively; numerical labels give the pooled p95 values.}
\label{fig:real_robot_commanded_jerk}
\end{figure}

\textbf{Results.}
Fig.~\ref{fig:real_robot_tasks} shows keyframe sequences from the physical tasks, Table~\ref{tab:real_robot_results} summarizes the corresponding success rates, and Fig.~\ref{fig:real_robot_commanded_jerk} compares the jerk of the TCP target trajectories actually issued to the robot. LieSpline-DP achieves higher observed success rates on all three tasks: 20/20 versus 2/20 trials on pouring, 20/20 versus 6/20 on bucket hooking, and 16/20 versus 11/20 on cube stacking. Across all tasks, LieSpline-DP reduces pooled p95 commanded jerk by over an order of magnitude compared with DP. Examples of real-robot execution are 
% provided in the supplimentary videos.
available on our \href{https://xieerxuan.github.io/LieSpline-DP/}{project webpage}.

In \emph{pouring}, abrupt baseline motions were frequently accompanied by splashing outside the target container. In some trials, the robot repeatedly switched between two poses while the water sloshed noticeably. We hypothesize that severe sloshing produced visual observations not represented in the demonstrations, destabilizing the policy's predictions. LieSpline-DP instead generated smooth, sustained pouring motions and succeeded in all 20 trials.

\begin{figure}[t]
\centering
\includegraphics[width=\columnwidth]{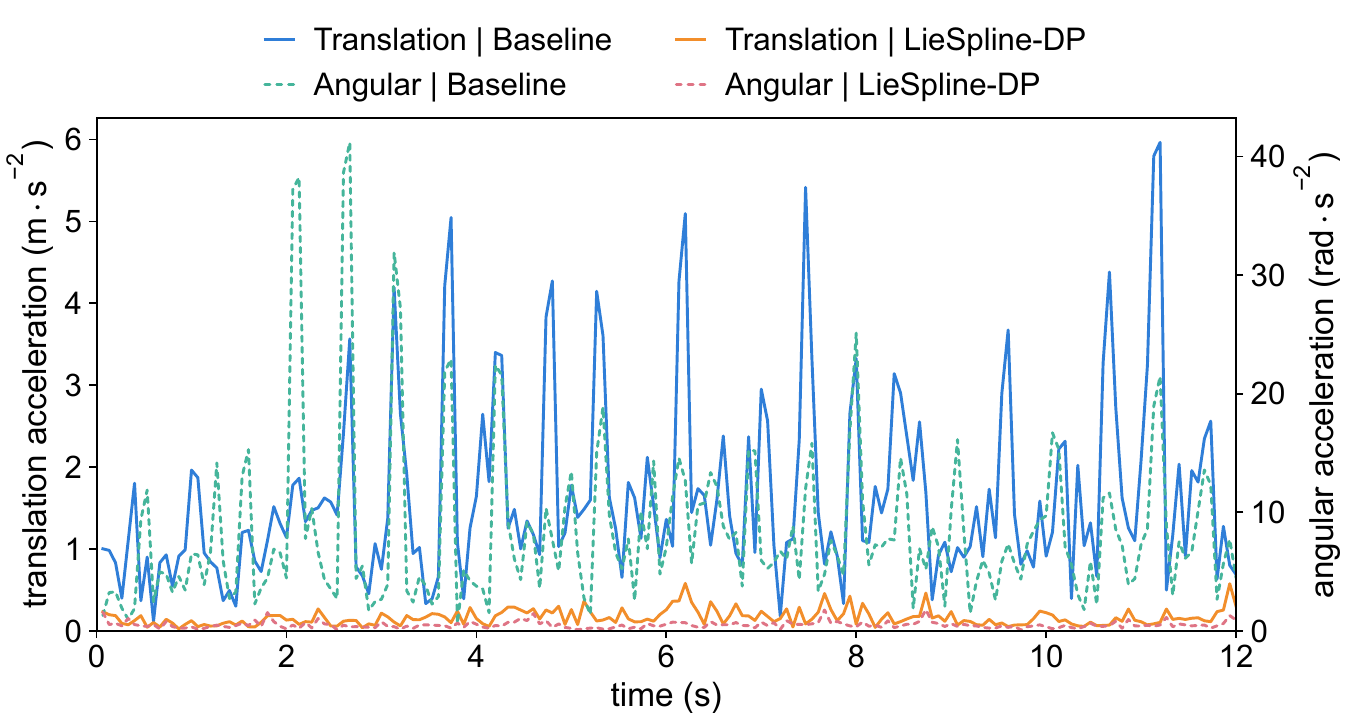}
\caption{\textbf{Commanded acceleration during bucket hooking.} Translational (left axis) and angular (right axis) acceleration magnitudes over the first 12~s of one rollout per method, estimated by finite differences of the $15$~Hz TCP target-pose sequences.}
\label{fig:hook_bucket_acceleration}
\end{figure}

In \emph{bucket hooking}, abrupt baseline motions were often accompanied by rope oscillations and subsequent recovery failures. By contrast, LieSpline-DP succeeded on the first hooking attempt in most trials. Even when the initial attempt failed, it recovered within a few additional attempts, with only modest hook oscillations rather than the pronounced swinging associated with baseline recovery failures. Fig.~\ref{fig:hook_bucket_acceleration} provides an illustrative time-series comparison: in these two rollouts, LieSpline-DP exhibits much smaller commanded acceleration peaks than DP.

In \emph{cube stacking}, baseline jitter sometimes led to unintended contacts that toppled a cube during approach or placement. Displaced cubes could fall outside the reachable workspace, preventing recovery. Both policies also occasionally hesitated before releasing the final cube until the episode timed out, although this behavior was less frequent with LieSpline-DP. We hypothesize that inheriting boundary control poses preserves part of the motion intent from the preceding execution segment, making LieSpline-DP less prone to stalling during replanning. The cube stacking results demonstrate that our spline-based policy is also capable of precise manipulation in real world. In an additional test of responsiveness, LieSpline-DP successfully recovers from 7 of 10 human perturbations that disturb a partial stack. This supplementary test evaluates LieSpline-DP only, without a DP comparison.

\subsection{Discussion}

Although LieSpline-DP achieves slightly lower overall success rates than the DP baseline in simulation, it produces substantially smoother trajectories. In contrast, on all three real-robot tasks, it achieves both higher success rates and smoother commanded motion. This contrast may partly reflect differences in task requirements: Robomimic Square and Tool Hang emphasize precise alignment and fine adjustments, which may be challenging to capture with a compact spline representation, whereas \textit{pouring} and \textit{bucket hooking} place greater demands on motion smoothness when interacting with liquids and flexible objects whose complex dynamics are sensitive to action jitter. To examine whether LieSpline-DP can also handle precision-demanding tasks on a physical robot, we also designed the \textit{cube stacking} experiment. Somewhat surprisingly, LieSpline-DP still outperforms the baseline. This advantage appears largely attributable to the mitigation of the two common baseline failure modes described above.

Overall, our observations suggest that simulation benchmarks may understate the real-world consequences of action jitter. LieSpline-DP's inherently smooth action representation supports smoother commanded motion and higher success rates across all three physical tasks.

\section{Limitations and Future Work}

Although representing actions as splines on $\mathrm{SE}(3)$ is, in principle, agnostic to the underlying action generator, this work evaluates it only with a Transformer-based diffusion policy. Future work could investigate this representation with other generative backbones, such as flow-matching policies~\cite{lipman2023flow,braun2024rfmp} and vision-language-action models~\cite{black2024pi0,kim2024openvla}, to assess whether the improvements in real-robot motion quality observed in this work generalize across model classes and broader task distributions. Beyond direct policy prediction, control-pose parameterization also provides a promising interface for planning as inference: instead of correcting a dense waypoint sequence, task costs and constraints can refine a compact set of control poses, while the local support and smooth decoding of B-splines propagate each correction over a bounded trajectory interval without sacrificing continuity~\cite{toussaint2009trajectory,carvalho2024mpd}. Integrating learned spline priors with inference-time optimization for collision avoidance, dynamic feasibility, and task-specific constraints is therefore a possible direction.

\section{Conclusion}
We presented LieSpline-DP, a Lie-group B-spline diffusion policy that represents end-effector actions as cumulative cubic B-splines on $\mathrm{SE}(3)$. By fixing boundary control poses from the previous plan and inpainting only future control poses, LieSpline-DP couples consecutive plans during asynchronous execution and guarantees cross-chunk $C^2$ continuity of the spline reference without post-hoc blending or refitting. Real-robot experiments demonstrate lower trajectory jerk and higher task success rates than the DP baseline, with particularly pronounced gains in tasks involving liquids and flexible objects.

\end{document}